\documentclass{IEEEtran4PSCC}
\usepackage{url}
\usepackage{bm}
\usepackage{algorithm2e}
\usepackage{hyperref}
\hypersetup{
colorlinks=true,
linkcolor=black,
filecolor=black,      
urlcolor=black,
citecolor=black
}
\usepackage[monochrome]{color}
\usepackage{amsmath,amsfonts}
\ifCLASSINFOpdf
   \usepackage[pdftex]{graphicx}
\else
   \usepackage[dvips]{graphicx}
\fi
\makeatletter
\let\old@ps@headings\ps@headings
\let\old@ps@IEEEtitlepagestyle\ps@IEEEtitlepagestyle
\def\psccfooter#1{%
    \def\ps@headings{%
        \old@ps@headings%
        \def\@oddfoot{\strut\hfill#1\hfill\strut}%
        \def\@evenfoot{\strut\hfill#1\hfill\strut}%
    }%
    \def\ps@IEEEtitlepagestyle{%
        \old@ps@IEEEtitlepagestyle%
        \def\@oddfoot{\strut\hfill#1\hfill\strut}%
        \def\@evenfoot{\strut\hfill#1\hfill\strut}%
    }%
    \ps@headings%
}
\makeatother

\psccfooter{%
        \parbox{\textwidth}{\hrulefill \\ \small{23rd Power Systems Computation Conference} \hfill \begin{minipage}{0.2\textwidth}\centering \vspace*{4pt} \includegraphics[scale=0.06]{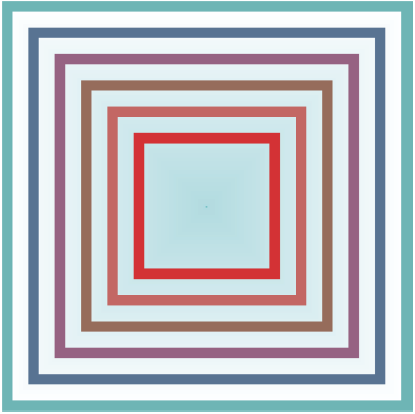}\\\small{PSCC 2024} \end{minipage} \hfill \small{Paris, France --- June 4 -- 7, 2024}}%
}

\begin{document}
\title{DiffPLF: A Conditional Diffusion Model for Probabilistic Forecasting of EV Charging Load}
\author{
\IEEEauthorblockN{Siyang Li, Hui Xiong, and Yize Chen}
\IEEEauthorblockA{Hong Kong University of Science and Technology (Guangzhou), Guangzhou
\\sli572@connect.hkust-gz.edu.cn, xionghui@hkust-gz.edu.cn, yizechen@ust.hk
}
}


\maketitle

\begin{abstract}
Due to the vast electric vehicle (EV) penetration to distribution grid, charging load forecasting is essential to promote charging station operation and demand-side management. However, the stochastic charging behaviors and associated exogenous factors render future charging load patterns quite volatile and hard to predict. Accordingly, we devise a novel \underline{Diff}usion model termed DiffPLF for \underline{P}robabilistic \underline{L}oad \underline{F}orecasting of EV charging, which can explicitly approximate the predictive load distribution conditioned on historical data and related covariates. Specifically, we leverage a denoising diffusion model, which can progressively convert the Gaussian prior to real time-series data by learning a reversal of the diffusion process. Besides, we couple such diffusion model with a cross-attention-based conditioning mechanism to execute conditional generation for possible charging demand profiles. We also propose a task-informed fine-tuning technique to better adapt DiffPLF to the probabilistic time-series forecasting task and acquire more accurate and reliable predicted intervals. Finally, we conduct multiple experiments to validate the superiority of DiffPLF to predict complex temporal patterns of erratic charging load and carry out controllable generation based on certain covariate. Results demonstrate that we can attain a notable rise of 39.58\% and 49.87\% on MAE and CRPS respectively compared to the conventional method.
\end{abstract}


\section{Introduction}
Rapid advancements of energy storage, fast-charging infrastructure, and carbon-reduction blueprints \cite{8930619} facilitate recent proliferation of electric vehicles (EVs). For instance, there will be 30-42 million EVs supported by 26-35 million charging piles in United States by 2030 \cite{nrel}. Such substantial EV penetration exerts additional large-scale, highly stochastic load to power networks, imposing unprecedented challenges on grid operation \cite{9808315}. EV charging load forecasting is crucial to host such vast charging demand, which can benefit operators of both distribution network and charging stations. For instance, grid operators are able to design the optimal coordinated dispatch of EVs and renewables in light of predicted charging power outcomes \cite{8521585}. Station managers can reduce the electricity procurement cost and optimize the real-time charging scheduling aided by the demand forecast information \cite{10121507}.

Point forecast is a classical way to procure future charging demand. In \cite{9078831}, a stacked recurrent neural network is forged by reinforcement learning to execute demand prediction for different charging scenarios. In \cite{ARIAS2016327}, charging demand is estimated by the joint statistical analysis for battery behaviors, traffic flow and weather data. However, since the real-world charging demand is extremely stochastic and volatile due to a set of uncertain factors (e.g. battery dynamics, traveller charging patterns and weather conditions), single deterministic charging load forecasts can not give station and power grid operators the most effective tools for load management and operations. Inaccurate prediction can raise operational costs and jeopardize power quality \cite{10121507}. Besides, it is more crucial for stakeholders to gain a group of reliable forecasts which are beneficial for stochastic optimization and robust decision making \cite{WU201755}. To this end, probabilistic forecasting is a promising approach to model the forecast uncertainties, which can be achieved by generating a host of plausible charging load profiles. Operators can exploit such probabilistic predictions to lessen charging energy deviation costs \cite{10121507}. For instance, \cite{phipps2023customized} proposes to forecast the uncertainty of EV parking duration.

In this work, we are interested in probing an accurate and reliable probabilistic forecasting model to tackle unknown charging load uncertainties. Forecasting both the values and uncertainties of EV charging scenarios is an emergent topic, while several works focus on predicting probable charging demand in urban areas. Quantile regression like~\cite{9790082} and~\cite{BUZNA2021116337} is a typical way to explicitly construct the prediction interval (PI) by learning a group of quantiles of different degrees. They utilize disparate neural networks to learn the relationships between historical charging data and these quantiles supervised by certain PI evaluation metrics. Nevertheless, such approach falls short in modeling complex temporal EV charging processes which are further complicated by conditional information such as weather and user behaviors. Another feasible approach is to directly quantify the point forecast uncertainties, which stem from both predictive model misspecifications and charging mode variability. For example, hidden state variations in the LSTM-based point predictor are captured by proximal policy optimization in \cite{9760076}. A novel queuing model is proposed in~\cite{9055130} to link mobile EV charging load to time-varying traffic flow, while a meta-learning method is introduced in~\cite{MetaProbformer} to address the data scarcity issue. 
However, these methods depend on both well-trained deterministic prediction model and assumption of the forecast error distribution. 

Essentially, the principal objective of probabilistic charging load forecasting is to obtain the predictive distribution of future charging demand profiles based on observed information. The currently heated generative models \cite{dhariwal2021diffusion} are able to approximate the complex distribution of high-dimensional data and generate various samples of high quality, which are promising to explicitly model the predictive distribution of charging time-series and yield a host of plausible future trajectories. Similar work has been done on the generic time-series analysis, like multivariate time-series forecasting \cite{pmlr-v139-rasul21a}, \cite{rasul2020tempflow} and imputation \cite{NEURIPS2021_cfe8504b}, all of which leverage generative models to estimate the target conditional distribution. However, these work do not explicitly devise a conditioning scheme to entangle the predicted charging load with historical data and relevant covariates.

In this paper, we aim to develop a data-driven generative model to directly learn the joint distribution of future charging load, which can be also conditioned on historical load data and associated covariates including weather forecasts, calendar variables and EV number. These informative covariates are non-negligible to achieve high-quality PIs. More importantly, EV charging station operators and grid operators are also interested in analyzing the impacts of certain variables on EV charging sessions~\cite{lee2020exploring}. We adopt the denoising diffusion model proposed in \cite{NEURIPS2020_4c5bcfec}, which has demonstrated expressive capacity to produce diverse high-fidelity images and perform controllable generation guided by text prompts \cite{9878449}. Diffusion models can transform the prior Gaussian noise to target charging load time-series by learning a parameterized reversal of diffusion process. Besides, diffusion models are quite efficient in training and inference, which also evade the mode collapse and training instability issues in other generative model such as generative adversarial networks (GANs) \cite{9765318}.

For the probabilistic forecasting of EV charging load task, we design a specific conditional denoising diffusion model entitled DiffPLF, to generate an array of plausible charging load profiles given historical charging demand and a group of informative covariates. We also incorporate the cross-attention mechanism~\cite{9878449} into the denoising network, which can condition each perturbed load time-series in the diffusion process on input conditional terms. 
In order to further gain more accurate and reliable probabilistic forecasts, our training objective is to make the 50\%-quantile (i.e. median) of produced outcomes be close to ground-truth charging load at each prediction step. In light of it, we propose a fine-tuning block over the pre-trained diffusion model via a 50\%-quantile deviation minimization (QDM) loss. Such QDM loss imposes a task-informed inductive bias on the diffusion model, and improves forecasting performance by around $40\%$ in contrast to standard quantile regression. We verify that DiffPLF can output more accurate predictive distribution and point forecast. Besides, it can be adapted to various prediction horizons and execute controllable generation conditioned on different EV numbers. Our code is publicly available at \href{https://github.com/LSY-Cython/DiffPLF}{https://github.com/LSY-Cython/DiffPLF} for better study of EV grid integration.
  


\section{Problem Formulation}
We start from describing the probabilistic charging load prediction task. We aim at capturing the forecast value along with uncertainties induced by stochastic and variable charging behaviors. The key is to explicitly model the conditional distribution of predicted charging load profiles given historical observations and correlated covariates. Specifically, at time $s$, we collect a series of past charging demand $p_{i}\in \mathbb{R}, i\in$ $[s-\omega,s]$, where $\omega$ is the length of the context window and $i$ is the time step.
We also prescribe a covariate set $\mathbf{r}$ which will be described shortly, and utilizing $\mathbf{r}$ is beneficial for accurate forecasts. Note that $\mathbf{r}$ shall be known in advance for the prediction horizon, which can be regarded as external information to constrain unwarranted forecasts. Based on the prior $\mathbf{p}\in \mathbb{R}^{\omega}$ and $\mathbf{r}$, we want to procure possible outcomes of future charging load $x^{j}_{0}\in \mathbb{R}, \; j\in [s+1,s+\tau]$, where $\tau$ stands for the length of the prediction horizon and we use subscript $0$ to align with the original data notation for diffusion models introduced in latter sections. Thereby, our goal is to approximate the conditional predictive distribution $q(\mathbf{x}_{0}|\mathbf{p},\mathbf{r})$ of future charging load profiles $\mathbf{x}_{0}\in \mathbb{R}^{\tau}$. The main challenge of this problem is how to derive diverse temporal patterns in $\mathbf{x}_{0}$ that are consistent with conditions $\mathbf{p}$ and $\mathbf{r}$.

\textbf{Covariates Selection}: Choosing informative covariates is critical for probabilistic time-series forecasting \cite{pmlr-v139-rasul21a}. In this paper, we find three types of covariates are most helpful: 1) \textit{Weather forecasts}. \textcolor{red}{Weather conditions are non-negligible for battery charging dynamics and EV travel behaviors, which can elicit various temporal modes of charging load \cite{9508419}. We utilize temperature forecast $\mathbf{u}\in \mathbb{R}^{\tau}$ and humidity forecast $\mathbf{v}\in \mathbb{R}^{\tau}$, which have been shown to be the most two influential weather factors for EV charging load \cite{ARIAS2016327}.} 2) \textit{Calendar variables}. EV mobility can be distinct both temporarily and spatially in terms of the day type (e.g. weekdays or weekends) \cite{NEAIMEH2015688}. In our setting, we use a one-hot vector $\mathbf{d}\in \mathbb{R}^{7}$ to signify seven days within a week. 3) \textit{EV number}. The total number of charged EV $e$ in the forecast window can affect both shape and peak value of predicted charging load time-series. We employ EV number as an unique condition to attest how its variation takes effect on ultimate forecasted profiles. In summary, the covariate set can be denoted as $\mathbf{r}=\{\mathbf{u},\mathbf{v},\mathbf{d},e\}$.

\section{Denoising Diffusion Models}
In this section, we explicate how the denoising diffusion model derive the complex distribution of charging load time-series, and how to extend it to perform conditional generation.

\begin{figure}[!t]
\centering
\includegraphics[width=0.5\textwidth]{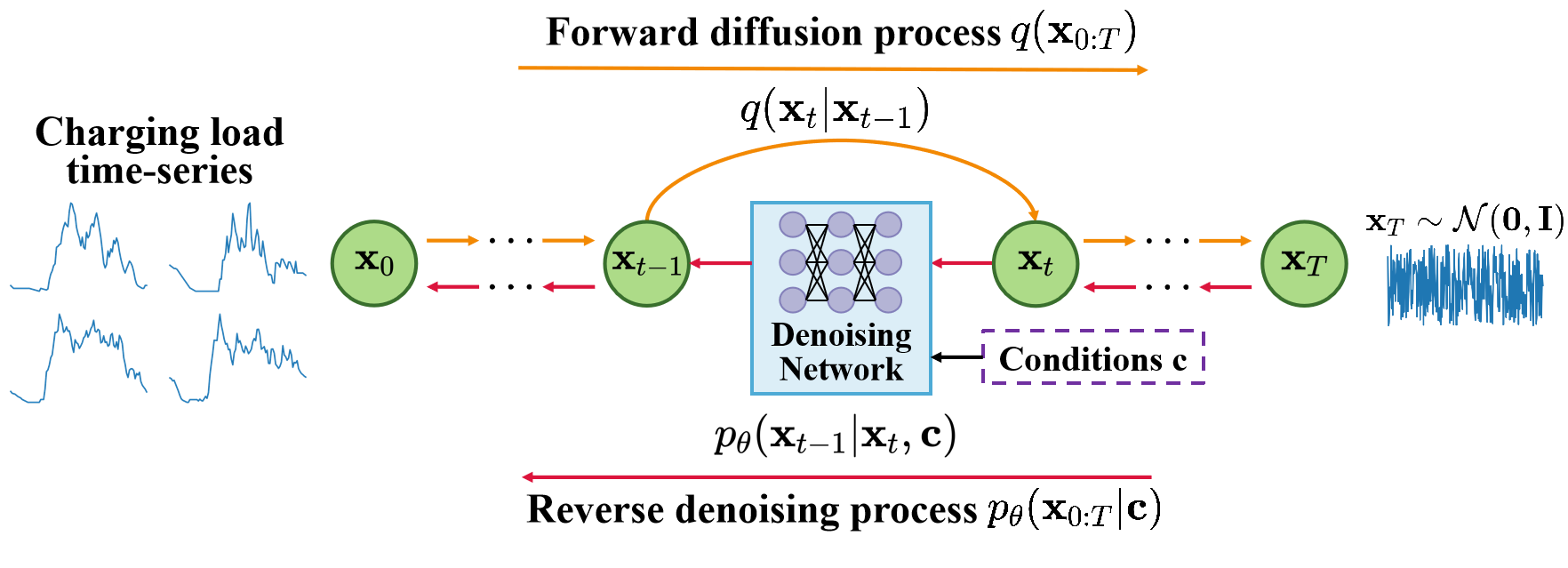}
\caption{The diagram of the conditional diffusion model.}
\vspace{-10pt}
\label{diffusion}
\end{figure}

Due to the eminent high-resolution synthesis capacity and efficient training advantage, denoising diffusion models have been actively applied to various multi-modal content creation fields~\cite{9878449}. The fundamental principle is 
inspired by the non-equilibrium thermodynamics, which implies that it is feasible to restore the true data distribution by simulating a physically reversible diffusion process~\cite{song2020score}. Firstly, we portray the \textit{forward diffusion process}, where real charging load profiles $\mathbf{x}_{0}$ are progressively transformed into standard Gaussian noise $\mathbf{x}_{T}$ after $T$-step diffusion procedures in total. This forward process is fixed to a Markov chain $q(\mathbf{x}_{1:T}|\mathbf{x}_{0})=\prod_{t=1}^{T}q(\mathbf{x}_{t}|\mathbf{x}_{t-1})$, which suggests step-wise Gaussian noise is gradually imposed on original $\mathbf{x}_{0}$. The forward transition $q(\mathbf{x}_{t}|\mathbf{x}_{t-1})$ has the form of Gaussian distribution as follows:
\begin{equation}
\centering
q(\mathbf{x}_{t}|\mathbf{x}_{t-1})=\mathcal{N}(\mathbf{x}_{t};\sqrt{1-\beta _{t}}\mathbf{x}_{t-1},\beta_{t}\mathbf{I}),
\label{eq:1}
\end{equation}
where $\beta_{t}$ stands for the magnitude of Gaussian noise added at each diffusion step $t$, and can be determined by the variance scheduling scheme in~\cite{NEURIPS2021_cfe8504b}. $\mathbf{x}_{t}$ denotes the perturbed charging load profile by the Gaussian noise with variance $\beta_{t}$. A critical property~\cite{NEURIPS2020_4c5bcfec} of this forward process is that we can obtain arbitrary $\mathbf{x}_{t}$ based on $\mathbf{x}_{0}$ in closed form:
\begin{subequations}
\label{eq:2}
\begin{align}
&q(\mathbf{x}_{t}|\mathbf{x}_{0})=\mathcal{N}(\mathbf{x}_{t};\sqrt{\alpha_{t}}\mathbf{x}_{0},(1-\alpha_{t})\mathbf{I}); \\
&\mathbf{x}_{t}=\sqrt{\alpha_{t}}\mathbf{x}_{0}+\sqrt{1-\alpha_{t}}\bm{\epsilon}, \bm{\epsilon}\sim\mathcal{N}(\mathbf{0},\mathbf{I}),
\end{align}
\end{subequations}
where $\alpha_{t}=\prod_{s=1}^{t}(1-\beta_{s})$. This helps significantly accelerate the forward noising procedure and improve the training efficiency of diffusion models.

Once the forward diffusion process is fixed, we can restore the genuine distribution of charging demand time-series $q(\mathbf{x}_{0})$ from the standard Gaussian $\mathbf{x}_{T}$ by learning a \textit{reverse denoising process}. Such reverse process can be defined as a learnable Markov chain $p_{\theta}(\mathbf{x}_{0:T-1}|\mathbf{x}_{T})=\prod_{t=1}^{T} p_{\theta}(\mathbf{x}_{t-1}|\mathbf{x}_{t})$ parameterized by $\theta$. In light of the physical properties of the invertible diffusion process \cite{NEURIPS2020_4c5bcfec}, if $\beta_{t}$ is small enough, the diffusion process is continuous and the reverse transition $p_{\theta}(\mathbf{x}_{t-1}|\mathbf{x}_{t})$ will hold the same function form as the forward transition $q(\mathbf{x}_{t}|\mathbf{x}_{t-1})$. Hence, we acquire a Gaussian reverse transition:
\begin{equation}
\centering
p_{\theta}(\mathbf{x}_{t-1}|\mathbf{x}_{t})=\mathcal{N}(\mathbf{x}_{t-1}; \bm{\mu} _{\theta}(\mathbf{x}_{t},t),\mathbf{\Sigma}_{\theta}(\mathbf{x}_{t},t)).
\label{eq:3}
\end{equation}
The main target of \eqref{eq:3} is to eliminate the noise added at step $t$ from the perturbed $\mathbf{x}_{t}$. Once the parameters $\theta$ in \eqref{eq:3} are determined, we can transform the prior Gaussian $\mathbf{x}_{T}$ into initial charging load data $\mathbf{x}_{0}$ through this reverse process.

We utilize the maximum likelihood estimation to learn the parameterized reverse transition in \eqref{eq:3} and approximate the real charging load distribution $q(\mathbf{x}_{0})$. We opt to minimize the negative log-likelihood of real demand data $-\log p_{\theta}(\mathbf{x}_{0})$ via its variational upper bound below:
\begin{equation}
\centering
-\log p_{\theta}(\mathbf{x}_{0})\le \mathbb{E}_{q(\mathbf{x}_{1:T}|\mathbf{x}_{0})}[-\log \frac{p_{\theta}(\mathbf{x}_{0:T})}{q(\mathbf{x}_{1:T}|\mathbf{x}_{0})}].
\label{eq:4}
\end{equation}
At each diffusion step $t$, we use a denoising network  to predict the noise added on $\mathbf{x}_{0}$. By decomposing \eqref{eq:4} into $T+1$ closed-form items, we get the following training objective for $\bm{\epsilon}_{\theta}$, while we refer the detailed derivations to \cite{NEURIPS2020_4c5bcfec}:
\begin{equation}
\label{eq:5}
\mathcal{L}_{t-1}=\mathbb{E}_{\mathbf{x}_{0},\bm{\epsilon},t}[\left\| \bm{\epsilon}-\bm{\epsilon}_{\theta}(\sqrt{\alpha_{t}}\mathbf{x}_{0}+\sqrt{1-\alpha_{t}} \bm{\epsilon},t)\right\|_2^{2}].
\end{equation}

Until now, we mainly cover generating EV charging load time-series $\mathbf{x}_0$ without any restrictions. In practice, many side features are collected together with charging load profiles, such as weather and past charging demand data, which help us better inform the generated curves. We adopt such features as conditional information, and develop the conditional diffusion model depicted in Fig. \ref{diffusion}. A straightforward way to achieve conditional generation is to update the reverse transition in \eqref{eq:3} into the conditional normal distribution below~\cite{9878449}:
\begin{equation}
\centering
p_{\theta}(\mathbf{x}_{t-1}|\mathbf{x}_{t},\mathbf{c})=\mathcal{N}(\mathbf{x}_{t-1}; \bm{\mu} _{\theta}(\mathbf{x}_{t},\mathbf{c},t),\tilde{\beta}_{t}\mathbf{I});
\label{eq:6}
\end{equation}
where $\mathbf{c}$ represents generic condition terms and in our problem setup, $\mathbf{c}$ consists of historical charging load and covariate set. For simplicity, $\mathbf{\Sigma}_{\theta}$ in \eqref{eq:3} is fixed to $\tilde{\beta}_{t}$ and $\tilde{\beta}_{t}=\frac{1-\alpha_{t-1}}{1-\alpha_{t}}{\beta}_{t}$ \cite{NEURIPS2020_4c5bcfec}, which alleviates the burden to learn $\mathbf{\Sigma}_{\theta}$ separately. Then, we can naturally condition the training objective in \eqref{eq:5} on $\mathbf{c}$:
\begin{equation}
\centering
\mathcal{L}_{t-1}=\mathbb{E}_{\mathbf{x}_{0},\mathbf{c},\bm{\epsilon},t}[\left\| \bm{\epsilon}-\bm{\epsilon}_{\theta}(\sqrt{\alpha_{t}}\mathbf{x}_{0}+\sqrt{1-\alpha_{t}} \bm{\epsilon},\mathbf{c},t)\right\|_2^{2}].
\label{eq:7}
\end{equation}

Once we train a conditional denoising network $\bm{\epsilon}_{\theta}$, we can sample $\mathbf{x}_{t-1}$ via the step-wise denoising operation below:
\begin{equation}
\label{eq:8}
\mathbf{x}_{t-1}=\frac{1}{\sqrt{1-\beta_{t}}}(\mathbf{x}_{t}-\frac{\beta_{t}}{\sqrt{1-\alpha_{t}}}\bm{\epsilon}_{\theta }(\mathbf{x}_{t},\mathbf{c},t))+\sqrt{\tilde{\beta}_{t}}\mathbf{z},
\end{equation}
where $\mathbf{z}\sim\mathcal{N}(\mathbf{0},\mathbf{I})$. This is actually a stochastic sampling procedure, because at each step $t$ in the reverse process, we also need to sample Gaussian noise $\mathbf{z}$.

\section{Probabilistic Forecasting Framework}
In this section, 
we detail how to construct DiffPLF by appling the diffusion model for probabilistic charging load forecasts, which includes a denoising network with cross-attention conditioning mechanism and task-specific fine-tuning. Holistic implementation of DiffPLF is depicted in Algorithm \ref{algo1}.

\begin{figure}[t]
\centering
\includegraphics[width=0.5\textwidth]{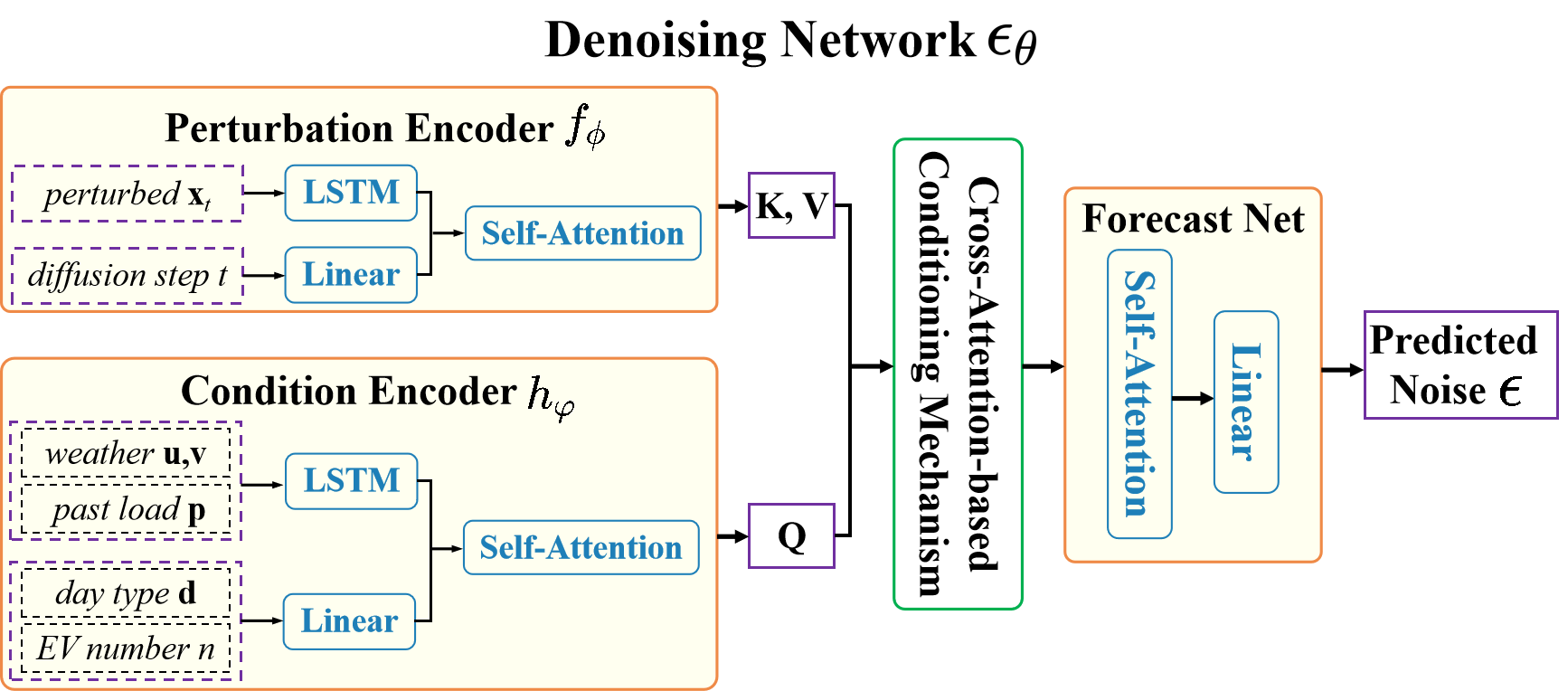}
\caption{The architecture of the proposed denoising network dedicated for the conditional diffusion model.}
\label{network}
\end{figure}

\subsection{Denoising Network}
To model the target conditional distribution $q(\mathbf{x}_{0}|\mathbf{p},\mathbf{r})$, we should train an effective denoising network $\bm{\epsilon}_{\theta }(\mathbf{x}_{t},\mathbf{p},\mathbf{r},t)$ by $\mathcal{L}_{t-1}$ in \eqref{eq:7}, whose input is the perturbed charging load profile $\mathbf{x}_{t}$, diffusion step $t$ and conditional terms $\mathbf{c}=\{\mathbf{p},\mathbf{r} \}$, and output is the noise $\bm{\epsilon}$ added on $\mathbf{x}_{t}$. How to incorporate the supplemental $\mathbf{c}$ into $\bm{\epsilon}_{\theta }$, i.e., to condition $\bm{\epsilon}_{\theta }$ on $\mathbf{c}$ is a vital issue. For instance, TimeGrad \cite{pmlr-v139-rasul21a} simply concatenates $\mathbf{x}_{t}$ and $\mathbf{c}$ into joint input vectors for LSTM units and does not explicitly use any conditioning schemes. Whereas in text-to-image generation, several conditioning ways such as classifier-based guidance \cite{dhariwal2021diffusion}, classifier-free guidance \cite{ho2022classifier} and cross-attention mechanism \cite{9878449} are proposed to yield proper images which are highly aligned with text semantics. We employ the cross-attention mechanism to entangle $\mathbf{x}_{t}$ with $\mathbf{c}$. 
The cross-attention in our method aims to discover latent information of conditions $\mathbf{c}$ that are correlated with predicted profiles $\mathbf{x}_{0}$.

As shown in Fig. \ref{network}, the denoising network $\bm{\epsilon}_{\theta }$ contains four components: 1) \textit{Perturbation encoder} $f_{\phi  }$, which feeds perturbed time-series $\mathbf{x}_{t}$ along with $t$ to a LSTM and linear layer respectively, and then utilizes a self-attention module to integrate the latent features of $\mathbf{x}_{t}$ and $t$. It actually amounts to the unconditional noise prediction manner defined in \eqref{eq:5}. 2) \textit{Condition encoder} $h_{\varphi }$, which helps represent the condition set $\mathbf{c}=\{\mathbf{p}, \mathbf{u}, \mathbf{v}, \mathbf{d}, e\}$. 
We concatenate the temporal data $\{\mathbf{p}, \mathbf{u}, \mathbf{v}\}$, and use LSTM to characterize their time dependencies. The discrete calendar vector $\mathbf{d}$ and EV number $e$ are handled by a linear layer. Then we also employ self-attention to fuse their latent features. 3) \textit{Cross-attention mechanism}, which is to condition the latent encoding of $\mathbf{x}_{t}$ and $t$ based on conditions $\mathbf{c}$, and this module is conducive to capture the conditional predictive distribution by \eqref{eq:7}. Denote $\mathbf{Q}=h_{\varphi }(\mathbf{p}, \mathbf{r})\cdot \mathbf{W}^{Q}$, $\mathbf{K}=f_{\phi  }(\mathbf{x}_{t}, t)\cdot \mathbf{W}^{K}$, $\mathbf{V}=f_{\phi  }(\mathbf{x}_{t}, t)\cdot \mathbf{W}^{V}$, then the cross-attention mechanism is formulated as follows:
\begin{equation}
\centering
\label{eq:9}
Attention(\mathbf{Q},\mathbf{K},\mathbf{V})=softmax(\frac{\mathbf{Q}\mathbf{K}^{T}}{\sqrt{d} } )\mathbf{V},
\end{equation}
where $\mathbf{W}^{Q}$, $\mathbf{W}^{K}$, $\mathbf{W}^{V}$ are three linear transformation matrix weights to be optimized, and $d$ is their hidden dimension. 4) \textit{Forecast network}, which finally uses a self-attention module coupled with a linear projection to transform output features of cross-attention to the predicted noise $\bm{\epsilon}$ added on $\mathbf{x}_{t}$.

\begin{figure}[t]
\centering
\includegraphics[width=0.5\textwidth]{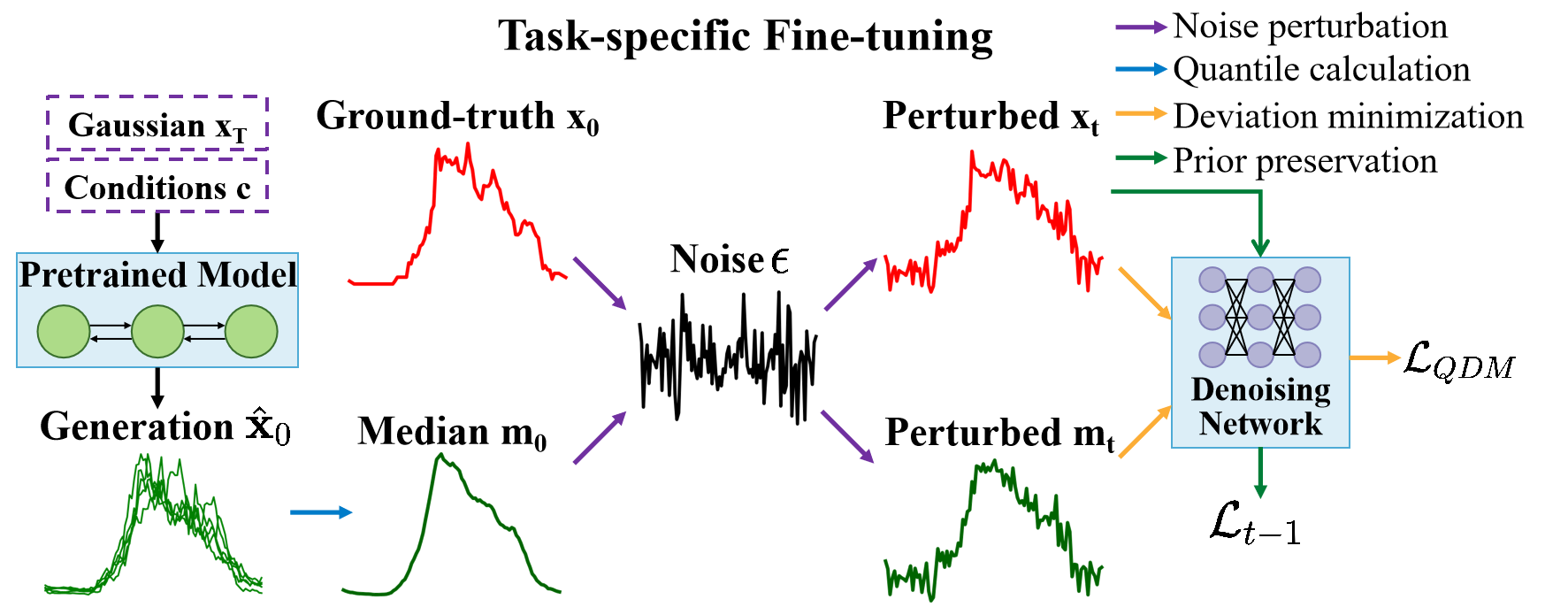}
\caption{The illustration of the fine-tuning procedure.}
\vspace{-10pt}
\label{refine}
\end{figure}

\subsection{Fine-tuning Technique}
\begin{figure}[!t]
\centering
\includegraphics[width=0.5\textwidth]{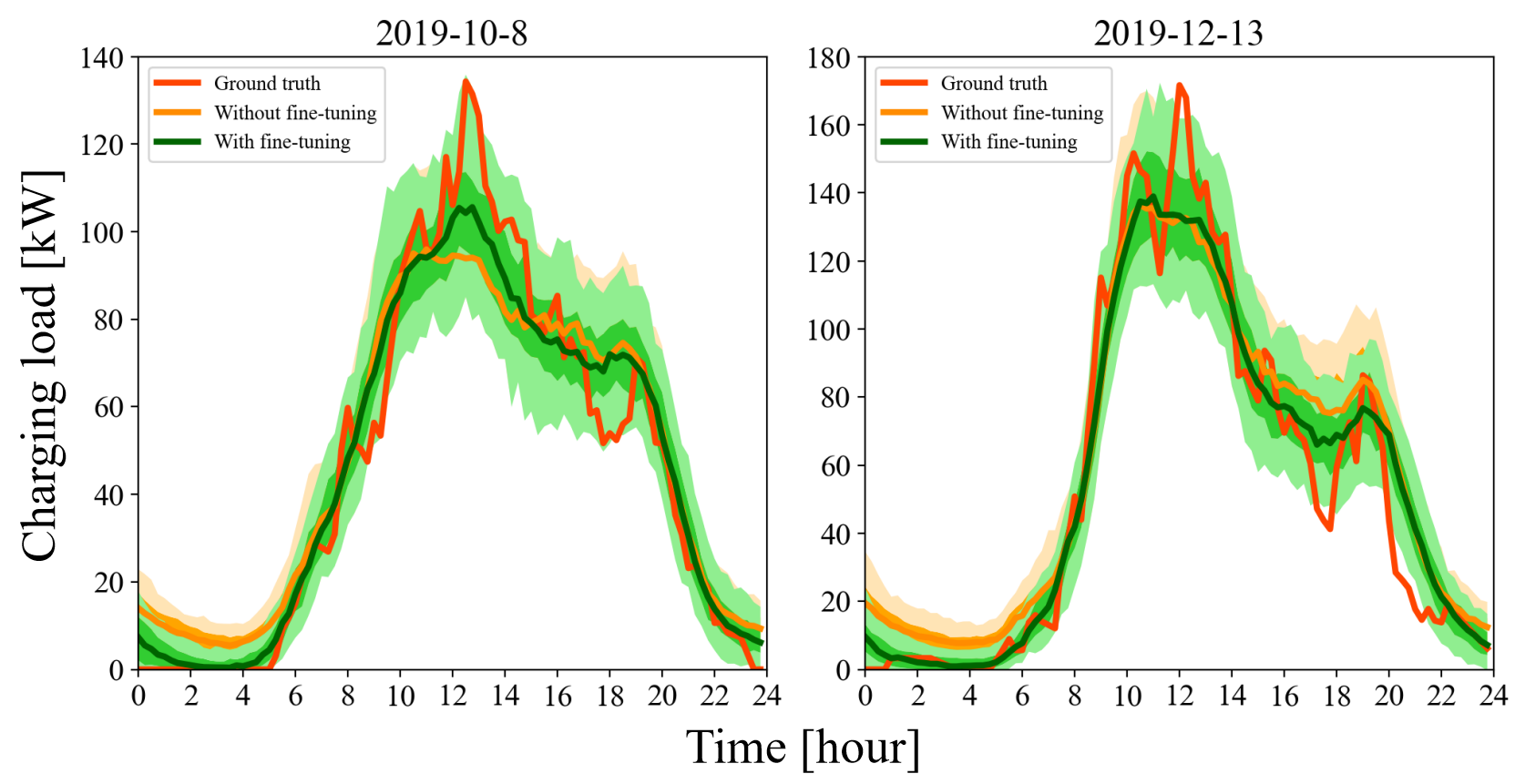}
\caption{Forecasting sample comparisons between the fine-tuned model and model without fine-tuning.}
\vspace{-10pt}
\label{compare PI}
\end{figure}

Indeed, we can train the former denoising network using \eqref{eq:7} and employ it to generate $N$ profiles $\left \{\hat{\mathbf{x}}^{n}_{0}\right \}^{N}_{n=1}$ via \eqref{eq:8}, where $\hat{\mathbf{x}}_{0}$ denotes the synthetic time-series. But these produced profiles may not form a high-quality PI \cite{pmlr-v139-rasul21a} that is required by probabilistic forecasting. To this end, we expect the 50\%-quantile $m_{0,i}$ of $N$ generated charging load $\left \{\hat{x}^{n}_{0,i}\right \}^{N}_{n=1}$ at each step $i$ should be close to the actual value $x_{0,i}$ as much as possible. In this work, we can treat $\mathbf{m}_{0} \in \mathbb{R}^\tau$ as point forecasts. Then our goal is to minimize the 50\%-quantile deviation $\left \| \mathbf{m}_{0}-\mathbf{x}_{0} \right \|^{2}_{2}$. 
Such term is an inductive bias and a \textit{task-specific refinement}. We leverage it to fine-tune the former model pre-trained by \eqref{eq:7}, where we explicitly enforce the conditional diffusion model to attain a particular objective. Besides, since $\mathbf{m}_{0}$ is constituted by the predicted outcomes of the pre-trained diffusion model, this fine-tuning procedure is realized \textit{by its own generated samples}. As shown in Fig. \ref{compare PI}, we can obtain sharper PI as well as more accurate deterministic forecasts after the task-informed fine-tuning stage.
As $\hat{\mathbf{x}}^{n}_{0}$ is generated by a stochastic sampling process which contains multiple times of iterations for $\bm{\epsilon}_{\theta }$, it is infeasible to directly use the divergence between $\mathbf{m}_{0}$ and $\mathbf{x}_{0}$ to refine the parameters of $\bm{\epsilon}_{\theta }$. To this end, we propose to leverage an alternative fine-tuning 50\%-quantile deviation minimization (QDM) objective:
\begin{equation}
\centering
\label{eq:10}
\mathcal{L}_{QDM}=\left \| \bm{\epsilon}_{\theta}(\mathbf{m}_{t},\mathbf{c},t)-\bm{\epsilon}_{\theta}(\mathbf{x}_{t},\mathbf{c},t) \right \|^{2}_{2},
\end{equation}
where $\mathbf{m}_{t}$ and $\mathbf{x}_{t}$ indicate we adpot (2b) to corrupt $\mathbf{m}_{0}$ and $\mathbf{x}_{0}$ by the same noise $\bm{\epsilon}$ at diffusion step $t$, and $c$ belongs to the ground-truth $\mathbf{x}_{0}$. Intuitively, minimizing $\left \| \mathbf{m}_{0}-\mathbf{x}_{0} \right \|^{2}_{2}$ can be equivalent to the goal of $\mathcal{L}_{QDM}$, which is also consistent with the training paradigm of the noise predictor. Moreover, we wish the conditional predictive distribution learned by the pre-trained diffusion model can not be impaired by $\mathcal{L}_{QDM}$, thus we also incorporate the $\bm{\epsilon}$-prediction loss of \eqref{eq:7} in this fine-tuning stage, then the total loss for $\bm{\epsilon}_{\theta }$ refinement is:
\begin{equation}
\centering
\label{eq:11}
\mathcal{L}_{ref}=\mathcal{L}_{t-1}+\lambda \mathcal{L}_{QDM},
\end{equation}
where $\lambda$ is the weight of QDM loss. The role $\mathcal{L}_{t-1}$ plays in $\mathcal{L}_{ref}$ can be comprehended as a \textit{prior preservation} item, which indicates that when the pre-trained diffusion model is carrying out the task-specific fine-tuning, it is able to retain its prior generative capability simultaneously. Our fine-tuning operation is illustrated in Fig. \ref{refine}.

\RestyleAlgo{ruled}
\begin{algorithm}
\caption{Implementation of DiffPLF}
\label{algo1}
\SetKwInput{KwInput}{Input}
\KwInput{training data $\mathcal{X}$, testing data $\mathcal{Y}$, diffusion step $T$}
\textbf{Stage 1}: Pretrain the denoising network $\bm{\epsilon}_{\theta }$ \\
\quad 1. Sample $t\sim \text{Uniform}(\{1,...,T\})$ \\
\quad 2. Sample $\mathbf{x}_{0}, \mathbf{c}\sim \mathcal{X}$, $\bm{\epsilon}\sim \mathcal{N}(\mathbf{0},\mathbf{I})$ \\
\quad 3. Optimize $\theta$ using $\mathcal{L}_{t-1}$ in \eqref{eq:7} \\
\textbf{Stage 2}: Task-specific fine-tuning \\
\quad 1. Sample $\mathbf{x}_{0}, \mathbf{c}\sim \mathcal{X}$ \\
\quad 2. Generate $\left \{\hat{\mathbf{x}}^{n}_{0}\right \}^{N}_{n=1}$ via \eqref{eq:8} and pretrained model \\
\quad 3. Calculate median $\mathbf{m}_{0}$ for generated $\left \{\hat{\mathbf{x}}^{n}_{0}\right \}^{N}_{n=1}$ \\
\quad 4. Obtain perturbed $\mathbf{x}_{t}$ and $\mathbf{m}_{t}$ using (2a) \\
\quad 5. Refine $\theta$ using $\mathcal{L}_{ref}$ in \eqref{eq:11} \\
\textbf{Stage 3}: Forecasting via inference \\
\quad 1. Sample $\mathbf{c}\sim \mathcal{Y}$, $\mathbf{x}_{T}\sim \mathcal{N}(\mathbf{0},\mathbf{I})$ \\
\quad 2. Predict $\mathbf{x}_{0}$ by iterating \eqref{eq:8} for $T$ times
\end{algorithm}

\section{Numerical Experiments}
\subsection{Experimental Setup}
1) \textit{Data description}: We harness a real-world dataset in the city of Palo Alto, California termed EV Charging Station Usage Open Data\footnote{\href{https://www.kaggle.com/datasets/venkatsairo4899/ev-charging-station-usage-of-california-city}{https://www.kaggle.com/datasets/venkatsairo4899/ev-charging-station-usage-of-california-city}}. 
It elaborates daily charging session details of individual stations, including charging durations and delivered energy for each EV. We resort to the transformation method in \cite{ARIAS2016327} to aggregate single battery charging curves to total charging load profiles with 15-min resolution. We fetch corresponding weather forecasts for Palo Alto from Meteostat platform\footnote{\href{https://dev.meteostat.net/}{https://dev.meteostat.net/}}. For all experiments involved in this section, we use historical EV charging demand and weather recording from 2016 to 2018 for both training and fine-tuning, while data in 2019 for testing.

2) \textit{Implementation details}: Regarding the historical data utilization, we exploit the aggregate charging demand of past 5 days to guide future load forecasts. In terms of the model architecture, we unify the hidden dimensions of LSTM, cross-attention and self-attention as 32, and the head number of two attention modules is set to 4. As for the noise scheduling, we follow the common quadratic scheme adopted in \cite{NEURIPS2021_cfe8504b}, where the start variance $\beta_{1}=0.0001$, the eventual variance $\beta_{T}=0.5$, and the number of diffusion steps $T$ is 200. 

We compile the whole DiffPLF using Pytorch library and implement it on a Linux service machine with a 48GB Nvidia A40 GPU. We use Adam optimizer to carry out the stochastic gradient descent with batch size of 16 for the noise predictor. During the pre-training and fine-tuning stage, the initial learning rate and total training epochs are 0.001/0.0002 and 200/100 respectively. Meanwhile, we find that we can achieve the best results when the weight of QDM loss $\lambda$ is 0.001 for model refinement. For every testing scenario, we randomly generate 1,000 possible trajectories of future charging load to constitute the target PI. \textcolor{red}{After the whole implementation of the proposed model, we find that the average training time of every epoch is 2.0174s and 2.5514s for the pre-training and fine-tuning stage respectively. Such discrepancy results from the fact that the denoising network will be executed only once during each standard diffusion training epoch, whilst being operated twice within each fine-tuning session. Besides, the mean inference time over each test case is 5.7511s, and note that when running DiffPLF on every test sample, we generate 1000 future charging load profiles in parallel.}

\begin{figure}[t]
\centering
\includegraphics[width=0.5\textwidth]{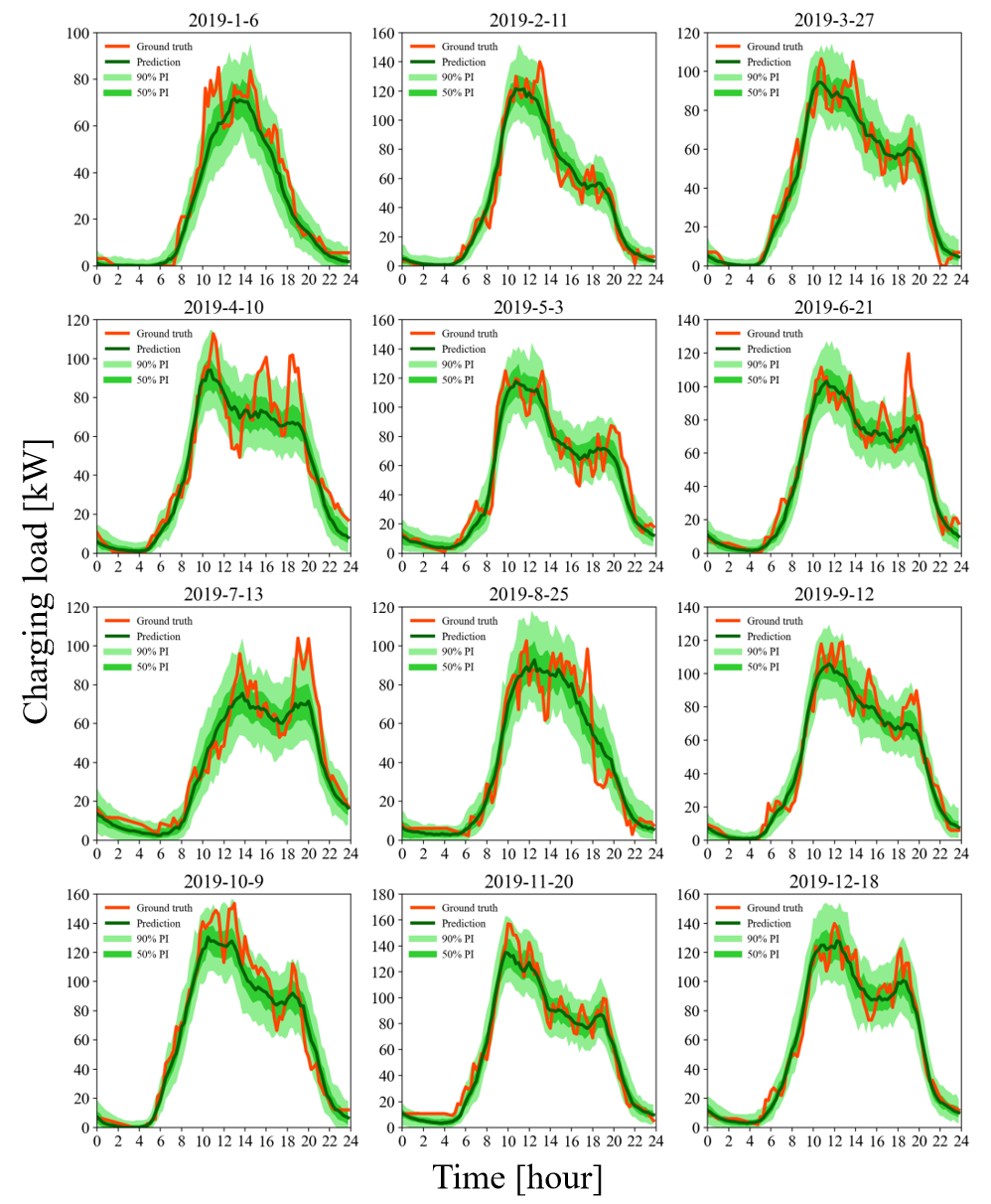}
\caption{Randomly selected testing samples. In each subplot, we depict the real charging load versus the generated day-ahead PI and point forecast.}
\vspace{-10pt}
\label{overall PI}
\end{figure}

\subsection{Simulation Results}
We focus on the day-ahead charging demand forecasting, where we predict charging load in the next day based on historical observations of past five days. We adopt two common metrics to evaluate the holistic performance of the proposed DiffPLF framework: 1) Mean Absolute Error (MAE)~\cite{MetaProbformer}. This indicator is aimed at assessing the point forecasting ability of the probabilistic forecasting model. We treat the median of the generated PI as the deterministic prediction. 2) Continuous Ranked Probability Score (CRPS)~\cite{pmlr-v139-rasul21a}. This index is utilized to judge the quality of the predictive distribution which is supposed to encompass the true realization. We compute the CRPS value for each prediction step and average that over all time steps as final CRPS for one testing sample. 

\begin{figure}[!t]
\centering
\includegraphics[width=0.5\textwidth]{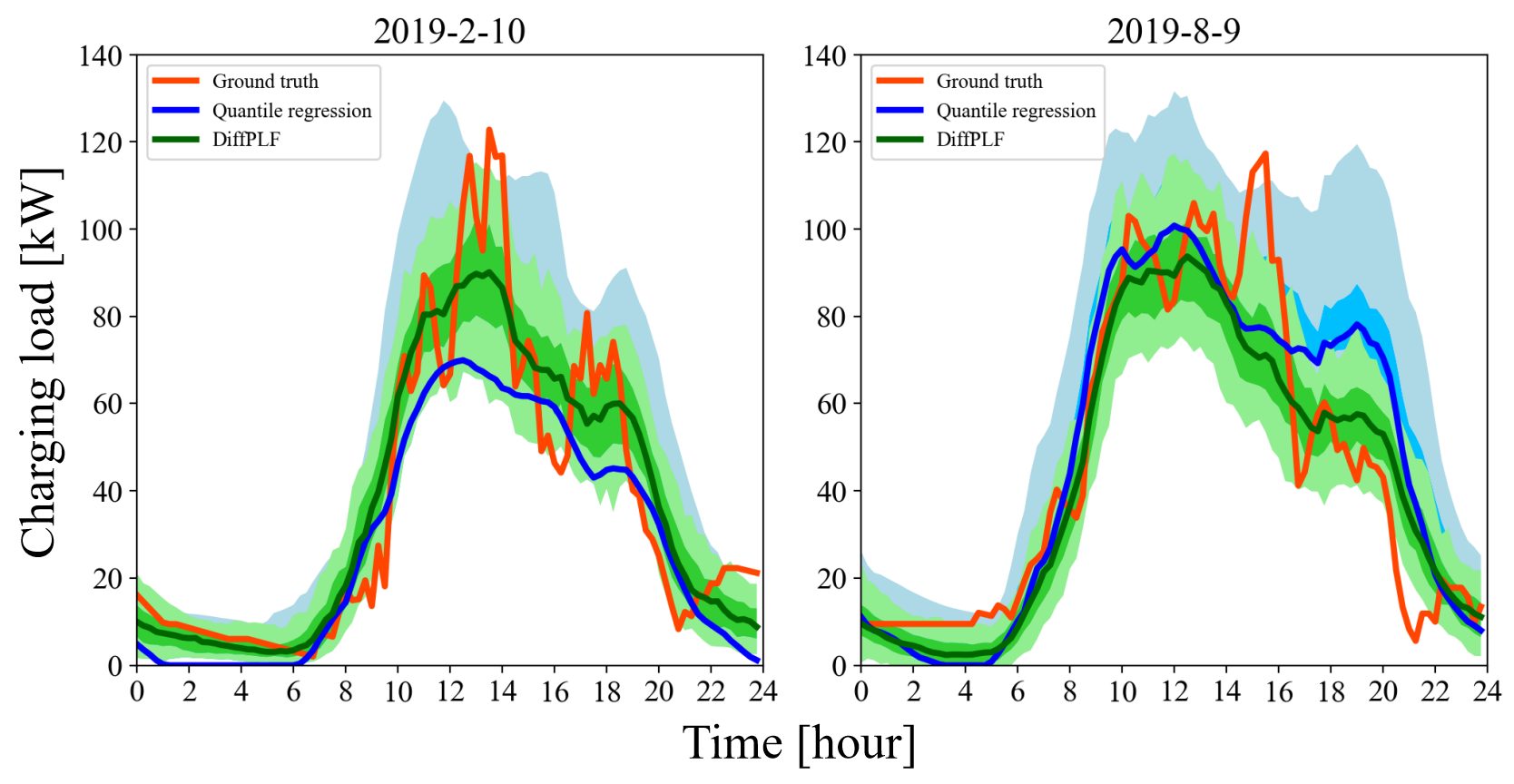}
\caption{Two forecasts of our generative DiffPLF versus quantile regression.}
\vspace{-10pt}
\label{beat PI}
\end{figure}

In Fig. \ref{overall PI}, we randomly draw and show 12 samples out of all testing samples in 2019. Evidently, the ground truth profile can not only be covered by shaded areas of either 50\% or 90\% PI, but also keep close to the produced point prediction to a large margin. It  validates that our DiffPLF is able to yield both accurate and reliable probabilistic forecasts. \textcolor{red}{To further verify the superiority of the diffusion-driven generative paradigm to model the predictive distribution, we alter the original training manner (i.e. noise prediction) of DiffPLF to quantile regression \cite{9790082}. Such method is trained to explicitly predict multiple probabilistic intervals and treated as a classical technique on probabilistic time-series forecasting. Specifically, we employ the developed noise predictor to directly estimate several quantiles (i.e. 5\%, 25\%, 75\%, 95\%) of future charging demand using the pinball loss \cite{JALALI2022108351} and exhibit its results in both the first row of Table \ref{overall metrics} and Fig. \ref{beat PI}.} We can observe that  DiffPLF can achieve considerably more accurate and sharper PI than the quantile regression method, because the extreme volatility of EV charging load renders it quite tough for the neural network to stably optimize the quantile loss. In Table \ref{overall metrics} we \textcolor{red}{summarize} the method comparisons, where w/o means model without the investigated component.

\begin{table}[!t]
\renewcommand{\arraystretch}{1.3}
\centering
\caption{Quantitative evaluation for different variants of DiffPLF.}
\label{overall metrics}
\begin{tabular}{ccc}
\hline
Method              & MAE          & CRPS                 \\ \hline
Quantile Regression & 11.852±3.936 & 10.107±2.124         \\
w/o covariates      & 7.842±2.065  & 5.592±1.570          \\
w/o cross-attention & 7.333±1.559  & 5.192±1.099          \\
w/o fine-tuning     & 7.227±1.489  & 5.111±1.041          \\
whole fine-tuning   & 7.200±1.607  & 5.089±1.122          \\
DiffPLF             & \textbf{7.161}±1.557  & \textbf{5.067}±1.094 \\ \hline
\end{tabular}
\end{table}

\textbf{Cross-attention Mechanism}: 
Previous works also bring up conditioning mechanisms like feature fusion in the latent space \cite{pmlr-v139-rasul21a}, \cite{NEURIPS2021_cfe8504b}.
For comparison, we utilize the element-wise addition to blend the latent embedding of both perturbation and condition encoder. From the third line in Table \ref{overall metrics}, a moderate drop more than 0.12 occurs on two metrics, indicating that cross-attention is a more effective way compared to latent fusion when modeling the conditional distribution of temporal charging load data.

\textbf{Supplementary Covariates}: 
Many previous methods fetch charging demand forecasts merely based on historical observations, ignoring the benefit of certain covariates which can be known for the prediction horizon in advance. Actually, the possible charging load distribution ought to comply with the covariate set, containing weather forecasts, day type and EV number, which are actually a kind of conducive constraints for the predictive model to produce more realistic and accurate forecasts. In order to investigate how DiffPLF benefits from such additional \textcolor{red}{covariates}, we purposely discard the input set $\mathbf{r}$ for the condition encoder and write its testing results in the second row of Table \ref{overall metrics}. After the covariate set is eradicated, we observe that there exists a salient decrease of 9.5\% and 10.4\% on MAE and CRPS respectively. It reflects that the appended covariates described in Section II are crucial for our approach to generate satisfactory probabilistic forecasts.

\textbf{Task-specific Fine-tuning}: In order to render the denoising diffusion method more amenable to the probabilistic time-series prediction task, we propose to fine-tune the pretrained diffusion model using the loss function $\mathcal{L}_{ref}$ defined in \eqref{eq:11}. $\mathcal{L}_{ref}$ consists of a prior preservation term $\mathcal{L}_{t-1}$ and a weighted item $\mathcal{L}_{QDM}$ to minimize the discrepancy between the median of generated PI and the ground-truth signal. To validate the efficacy of this fine-tuning trick, we compare the performance of entire DiffPLF and its pre-trained version without fine-tuning in Table \ref{overall metrics}. Apparently, DiffPLF outcomes degrade to a mild extent after removing its task-specific refinement, which indicates that the proposed fine-tuning technique is able to improve the effect of diffusion-based generative modeling on probabilistic forecasting for EV charging load.

Furthermore, we look into how different fine-tuning methods affect the final outcomes. In the last two rows of Table \ref{overall metrics}, we show that merely fine-tuning the former part of the noise predictor (while latter weights of cross-attention and parallel encoders are frozen) is better than refining the whole model. 
The performing gap between such two types of fine-tuning may arise from the implicit adversity of $\mathcal{L}_{QDM}$ for optimizing the latent encoding of heterogeneous input data and conditioning mechanism, which instead prefer to be optimized by the $\bm{\epsilon}$-prediction manner.

\begin{table}[!t]\color{red}
\renewcommand{\arraystretch}{1.3}
\centering
\caption{DiffPLF performance under different diffusion steps.}
\label{varying step}
\begin{tabular}{ccc}
\hline
Diffusion step T & MAE         & CRPS        \\ \hline
100              & 9.127±1.894 & 6.458±1.362 \\
150              & 7.313±1.555 & 5.177±1.106 \\
200              & 7.227±1.489  & 5.111±1.041 \\
250              & 9.569±2.096 & 6.780±1.492 \\
300              & 9.503±1.973 & 6.742±1.415 \\ \hline
\end{tabular}
\end{table}

\textcolor{red}{\textbf{Varying diffusion step $T$}: We investigate the model performance with respect to different settings of diffusion step $T$, since $T$ is one of the key factors that can determine the ultimate generation outcomes of discrete-time diffusion models \cite{nichol2021improved}. We conduct this sensitivity analysis just in the pre-training context, and forecasting outcomes of default $T=200$ is shown in the fourth row of Table \ref{overall metrics}, while evaluation results for other $T$ values are exhibited in Table \ref{varying step}. Totally, DiffPLF can achieve the best results on $T=200$ but be relatively less effective on other four $T$ settings. In light of \cite{nichol2021improved} and \cite{song2020score}, larger $T$ will give rise to heavy tails in the noise schedule which can deteriorate the learning efficiency of denoising network. Smaller $T$ can increase the discretization errors of continuous stochastic diffusion equations and render the Gaussian form presumption on the reverse transition \eqref{eq:3} less valid. How to determine the best $T$ for different forecasting scenarios more properly is of great significance, we leave it for future work.}

\begin{figure}[t]
\centering
\includegraphics[width=0.5\textwidth]{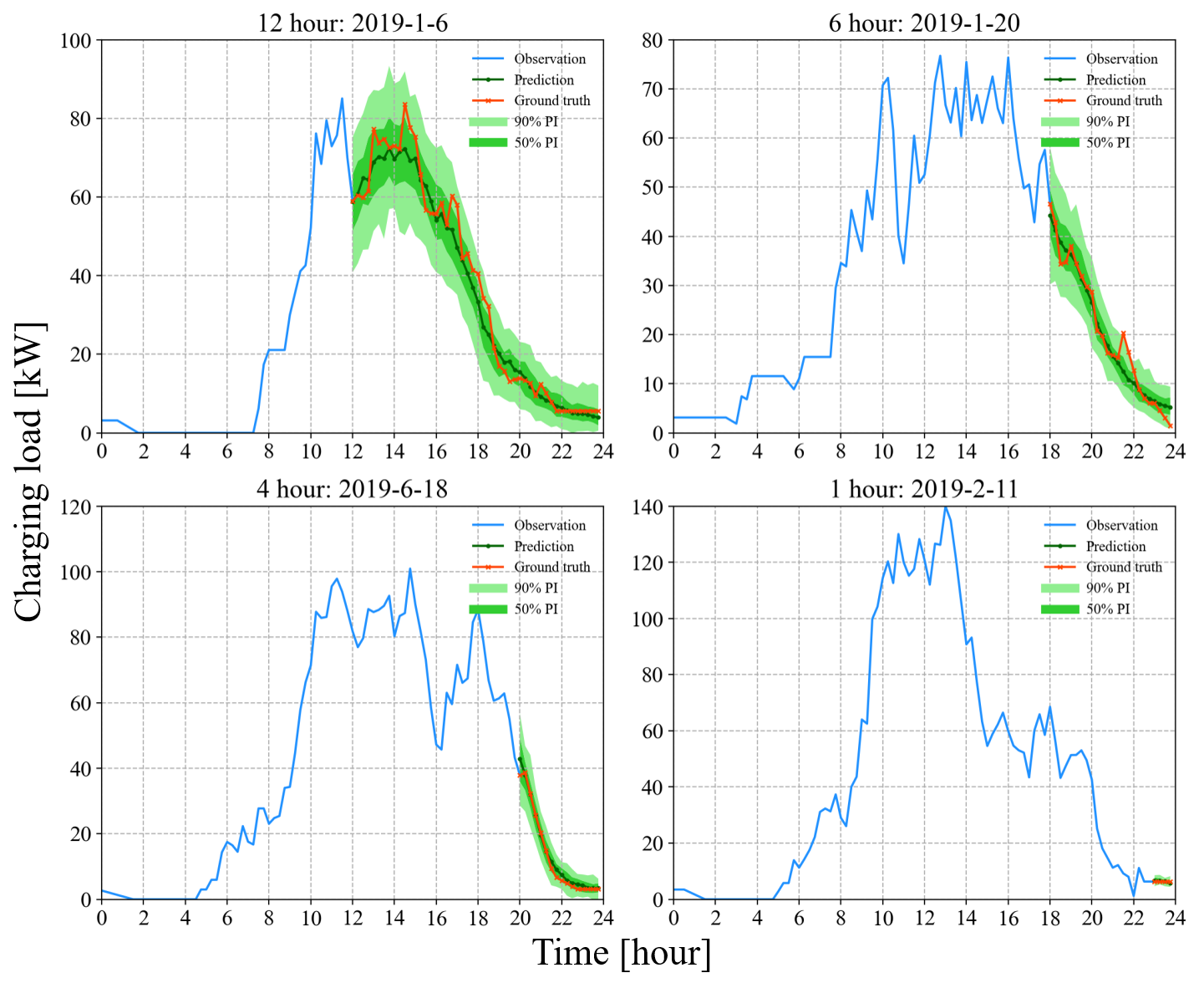}
\caption{Examples of produced PI with different prediction horizons.}
\vspace{-10pt}
\label{varying horizon}
\end{figure}

\begin{figure}[!t]
\centering
\includegraphics[width=0.5\textwidth]{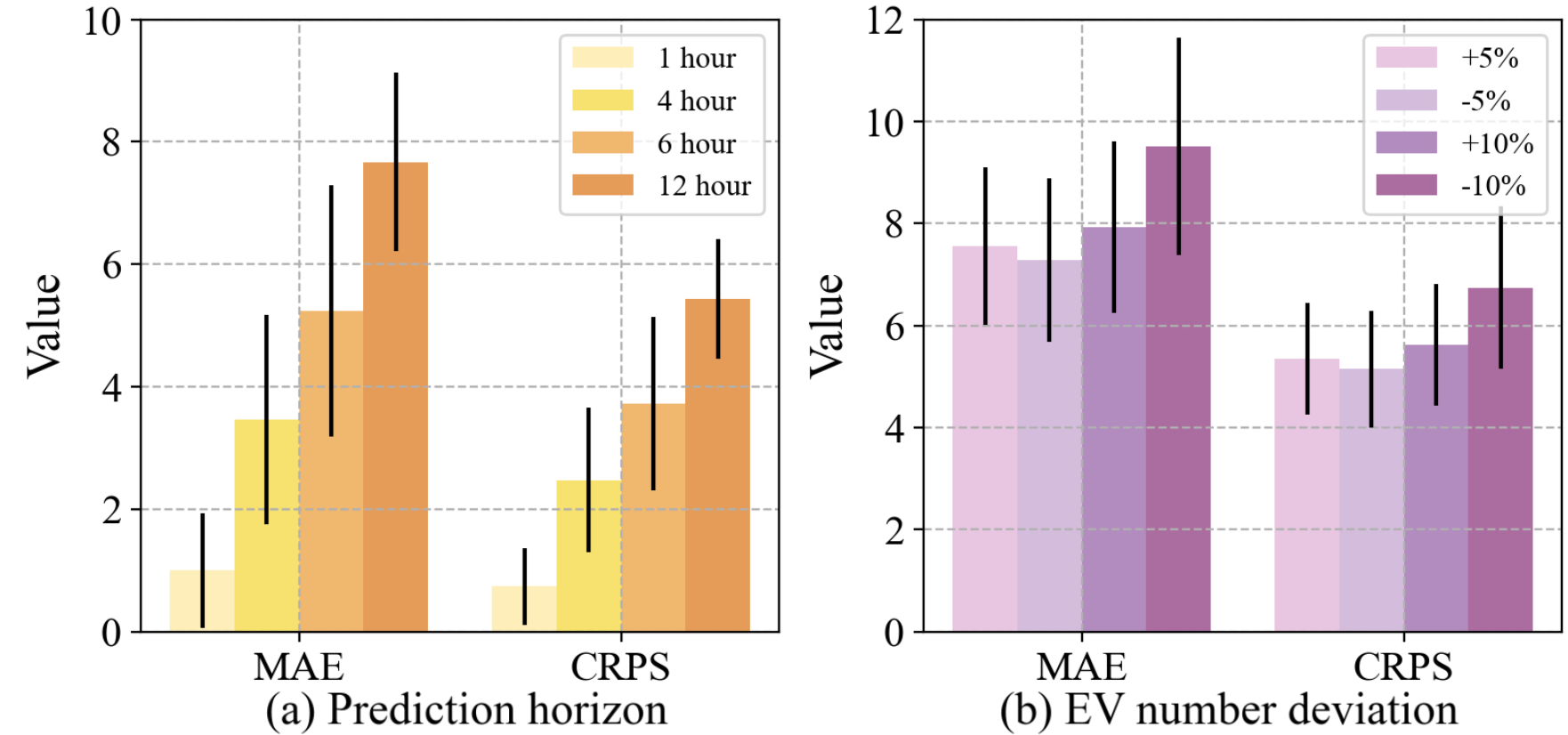}
\caption{Mean and variance of MAE and CRPS for two additional analysis: (a) Varying prediction horizons. (b) Different deviations of EV number.}
\vspace{-10pt}
\label{add metric}
\end{figure}

\textbf{Varying prediction horizons}: Here, we illustrate DiffPLF can be seamlessly scaled to different prediction horizons on top of precedent day-ahead forecasting. For the original 24h horizon $[s+1,s+\tau]$, we assume that charging demand in $[s+1,s+\eta],0<\eta<\tau$ has been measured and that in $[s+\eta+1,s+\tau]$ should be forecasted. Then, the prediction length is changed from $\tau$ to $\tau-\eta$. We can achieve this target by just modifying the input data for $\mathcal{L}_{QDM}$, without overriding the pre-trained diffusion model. Specifically, we simply fix $\mathbf{m}^{s+1:s+\eta}_{t}=\mathbf{x}^{s+1:s+\eta}_{t}$, leaving the output demand in $[s+\eta+1,s+\tau]$ to be determined. We consider a set of 12h, 6h, 4h and 1h forecasting horizons with examples shown in Fig. \ref{varying horizon}, where we can find that DiffPLF can also yield sharp and reliable PI for various prediction horizons. As is shown in Fig. \ref{add metric} (a),  the fine-tuned model holds consistent performance under varying forecasting lengths under both MAE and CRPS metrics. 

\begin{figure}[!t]
\centering
\includegraphics[width=0.5\textwidth]{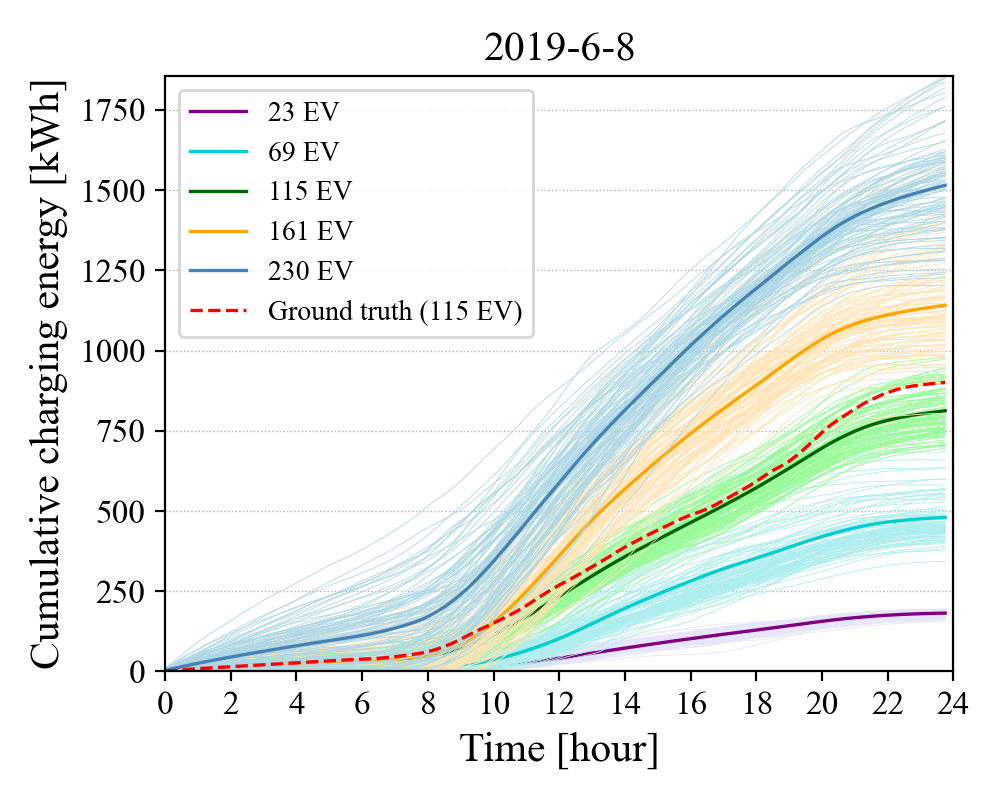}
\caption{Predicted daily cumulative charging energy profiles under various EV numbers. The bold line depicts the mean value of generated profiles (plotted by light lines) for each group with a fixed number of EVs.}
\label{EV number}
\end{figure}

\textbf{Varying EV numbers}: In practice, operators may be interested to analyze the charging load under different number of EV customers. To this end, we wish to investigate how the variable $e$ affects forecasting outcomes both visually and quantitatively. We fix the already trained DiffPLF and only adjust the input $e$ for the condition encoder. We select a testing sample on June 8th and give five different $e$, and the resulting profiles under each $e$ are depicted in Fig. \ref{EV number} respectively. Note that we showcase the cumulative charging energy, which are integration of instantaneous charging load. This shows DiffPLF can generate charging energy curves with unique magnitudes and ascending trends. It also suggests that our diffusion model is able to attain controllable generation conditioned on various EV numbers. Moreover, we also want to study the impact of EV quantity deviations on the accuracy of probabilistic forecasts. In Fig. \ref{add metric}(b), we exhibit numeric assessments when $e$ diverges from its ground-truth value by $\pm 5\%$ and $\pm 10\%$. We find that DiffPLF is robust to errors of EV number in total, except for the $-10\%$ group, where our model shows a modest drop on prediction results. \textcolor{red}{These results together verify that DiffPLF can be generalizable to EV numbers, forecasting horizons and data samples. In future work, it is also intriguing to benchmark diffusion model's efficacy across different patterns of EV charging datasets.}

\section{Conclusion and Outlook}
In this paper, we focus on forecasting EV charging load in a probabilistic way by proposing a novel conditional diffusion model DiffPLF. DiffPLF combines the denoising diffusion-driven generation and cross-attention mechanism to capture the predictive distribution conditioned on past demand and complementary covariates. A task-specific fine-tuning approach is devised to further ameliorate the quality of produced prediction intervals. Numerical experiments verify DiffPLF can achieve satisfactory probabilistic demand forecasting and controllable charging profile prediction under flexible look-ahead horizons. \textcolor{red}{Since the current method requires an additional task-informed fine-tuning operation to improve the prediction accuracy, we look forward to developing a more efficient end-to-end diffusion model which can be specialized in probabilistic load forecasting in future work. Besides, we hope to extend our model to predict long-term charging load which is beneficial for charging infrastructure planning. We also intend to explore multivariate diffusion-based generative model to handle EV charging and more general energy time-series.}

\bibliographystyle{IEEEtran}
\bibliography{reference}

\end{document}